\documentclass[11pt]{article}
\usepackage[margin=1in]{geometry}
\usepackage[T1]{fontenc}
\usepackage{lmodern}
\usepackage{microtype}
\usepackage{amsmath,amssymb,amsthm}
\usepackage{booktabs}
\usepackage{graphicx}
\usepackage{xcolor}
\usepackage{enumitem}
\usepackage{caption}
\usepackage{float}
\usepackage[hidelinks]{hyperref}
\usepackage{xurl}

\setlist{nosep,leftmargin=1.4em}
\newtheorem{proposition}{Proposition}

\newcommand{\vstar}{v^{*}}
\DeclareMathOperator*{\argmax}{arg\,max}

\title{Gumbel Substitution: Removing Sampler Drift from RL\\Without Trusting the Sampler}
\author{Romain Fanucci \qquad Malik El Hassani \\ Reliquary}
\date{November 2026}

\usepackage{array}
\usepackage{tikz}
\usetikzlibrary{positioning,arrows.meta}

\begin{document}
\maketitle

\begin{abstract}
When RL rollouts come from untrusted samplers on arbitrary hardware, precision and checkpoint lag, the sampling distribution $q$ differs from the trainer's policy $p$, and training on the sampler's tokens adds a drift term equal to supervised fine-tuning toward $q$. Published corrections (TIS, MIS, IcePop, score centering, GSPO, CISPO) all need $q(y)$, which a permissionless sampler reports and can misreport. We introduce Gumbel substitution. The coordinator fixes public Gumbel noise for every token position, bit-identical across GPUs, and the sampler draws $y=\argmax(\ell^q+G)$. The trainer recomputes $\vstar=\argmax(\ell^p+G)$, the token its own policy would have drawn with the same noise, and scores $\vstar$ in the sampler's context. For a sampler that follows the protocol, $\vstar$ is an exact sample from $p$ given the prefix, so the drift vanishes in expectation for any $q$, without sampler-reported probabilities. We measure the residual left by the deployed heads, bound the one added by admission gates, and state what an adaptive sampler can still do. The same comparison gives a per-token match rate that admits and pays samplers. In the lab, on a 9B model with the same head on both sides, the expected drift of uncorrected training is at least 170 to 4,800 times that of substitution. On a 2B model under strong mismatch, uncorrected training collapses while substitution stays stable, comparable to MIS with score centering given the true $q$. On a permissionless network, a 9B model was trained for 158 optimizer steps on 15.75B tokens from heterogeneous samplers, with no skipped step and 1.21\% of tokens substituted; it improved on all eight held-out benchmarks.
\end{abstract}

\section{Introduction}

\emph{Shared Gumbel noise lets a trainer cancel sampler drift without ever reading the sampler's probabilities.}

Rollout generation dominates the compute of RL for language models: about $4.5\times$ the training FLOPs in INTELLECT-2~\cite{intellect2}, and 80 to 84\% of the GPUs in recent large runs, which report inference-to-training ratios of 3.9:1 to 5.4:1~\cite{beam} and 4,096 inference chips out of 4,864~\cite{mai}. Moving it onto a permissionless network of independent GPUs would cut its cost, and recent work has made this practical: TOPLOC~\cite{toploc} makes untrusted inference verifiable, and INTELLECT-2~\cite{intellect2} trained a 32B model asynchronously on rollouts from untrusted machines. One problem remains open, and it is the one that destabilizes training. On such a network, samplers choose their hardware, their precision and the checkpoint they sample from, and the trainer accepts their tokens without trusting them. We deployed the method on Reliquary, subnet 81 of Bittensor, an open network in which a coordinator (the subnet's validator) pays anyone for compute it can verify; the samplers are independent operators paid per verified token.

The sampler's distribution $q$ then differs from the trainer's policy $p$. Trained on the sampler's tokens, the policy gradient contains a term that is supervised fine-tuning toward $q$ (Section~\ref{sec:problem}). Left alone, this drift destabilizes training: uncorrected GRPO~\cite{grpo} collapses on mixed-precision, mixed-accelerator samplers~\cite{modal}, and in our runs (Section~\ref{sec:minitrain}). The published corrections reweight, mask or center each token using $q(y)$~\cite{modal,yao,liu,icepop,centering}. On a permissionless network, $q(y)$ is reported by the party being paid, cannot be recomputed without the sampler's exact numerics, and the trainer does not know which checkpoint produced it.

Seed-synchronized sampling offers a way out. If the noise is fixed in advance, tokens become nearly deterministic given the model, which makes them auditable~\cite{difr}. We use the same noise for a second purpose: to rewrite the training target. At each position, the trainer computes the token $\vstar$ its own policy would have drawn with the sampler's noise, and scores $\vstar$ in the sampler's context. The sampler's tokens are never rewritten as context, only replaced as targets.

\paragraph{Contributions.}
\begin{enumerate}
\item \textbf{The substitution loss.} Given the prefix, $\vstar$ is an exact sample from $p$, so the drift term vanishes in expectation for any $q$, any hardware and any lag, for a sampler that follows the protocol and without sampler-reported probabilities (Section~\ref{sec:guaranteed}). We also state what it does not guarantee and the threat model it covers (Sections~\ref{sec:notguaranteed} and~\ref{sec:threat}).
\item \textbf{A deployable noise construction.} Per-checkpoint committed secrets, per-position hashing, and a Gumbel transform that is bit-identical on A100, RTX 4090, H100 and CPU. Gumbel-max disagrees 2.1 to $2.6\times$ less often than inverse-CDF sampling and stays within the worst-case guarantee of communication-free coupling~\cite{daliri} (Section~\ref{sec:gumbel-vs-icdf}).
\item \textbf{Controlled evidence.} Drift measured at a fixed checkpoint, staleness from 0 to 8 steps, a branching test of the cross-credit bias, and mini-trainings in which uncorrected training collapses and substitution does not. We also report the setting where substitution loses (Section~\ref{sec:minitrain}).
\item \textbf{Production evidence.} 158 optimizer steps on a 9B model over 11 days, 15.75B trained tokens from 161 sampler accounts across 7 environments, with no skipped or clipped step, match rate 0.979, 1.21\% substitution, per-sampler match-rate clusters that match the lab's precisions, and gains on all eight held-out benchmarks (Section~\ref{sec:prod}).
\end{enumerate}

\paragraph{Code.} The production coordinator, trainer, sampler and gates are open source at \url{https://github.com/reliquadotai/reliquary}.

\section{Problem}\label{sec:problem}

\emph{Training on another distribution's tokens adds a distillation term toward that distribution, and removing it has so far required knowing that distribution.}

\paragraph{Setting.} The trainer holds weights $\theta$, the last published checkpoint, with policy $p=p_\theta$. A sampler receives a prompt and samples a completion $y$ from its own distribution $q$. $q$ differs from $p$ through kernels, precision (for example fp8 weights or KV cache), hardware, and checkpoint lag. The sampler sends tokens only: no log-probability, no reward.

\paragraph{Drift.} With rollouts drawn from $q\neq p$, the REINFORCE / GRPO gradient at a position decomposes as~\cite{centering}:
\begin{equation}
\mathbb{E}_q\big[R\,\nabla \log p_y\big] \;=\; \mathbb{E}_q[R]\sum_v q_v\,\nabla \log p_v \;+\; \mathrm{Cov}_q\big(R,\ \nabla \log p_y\big).
\end{equation}
The second term is the learning signal. The first is a distillation toward the sampler: under $p$ it would vanish, since $\sum_v p_v \nabla\log p_v = 0$. With a constant advantage, the expected update on the logits is $\sum_v (q_v - p_v)\nabla \ell_v$. Recomputing $\pi_{\text{old}}$ on the trainer does not remove it: with one optimizer step per batch, the PPO~\cite{ppo} ratio is identically 1.

\paragraph{Why the known corrections do not apply.} TIS~\cite{yao}, MIS~\cite{liu}, IcePop~\cite{icepop}, score centering~\cite{centering}, GSPO~\cite{gspo} and CISPO~\cite{cispo} all use $q(y)$, and score centering needs $q$ on the top candidates. On a permissionless network these values come from the sampler. They cannot be recomputed without reproducing the sampler's numerics, and the coordinator does not know which checkpoint the sampler used. A reported $q$ is also an attack surface, since it sets the weight of the sampler's own tokens in the update.

\paragraph{Deployment constraints.} We require a correction that works fully asynchronously, lets samplers choose checkpoint and precision, verifies against the latest checkpoint only, and pays per verified token.

\section{Method}\label{sec:method}

\emph{The coordinator publishes the noise, the sampler draws with it, and the trainer scores its own pick under that noise in the sampler's context.} The sampler's tokens only ever serve as context. Every training target is the trainer's own pick under noise fixed before the sampler sampled.

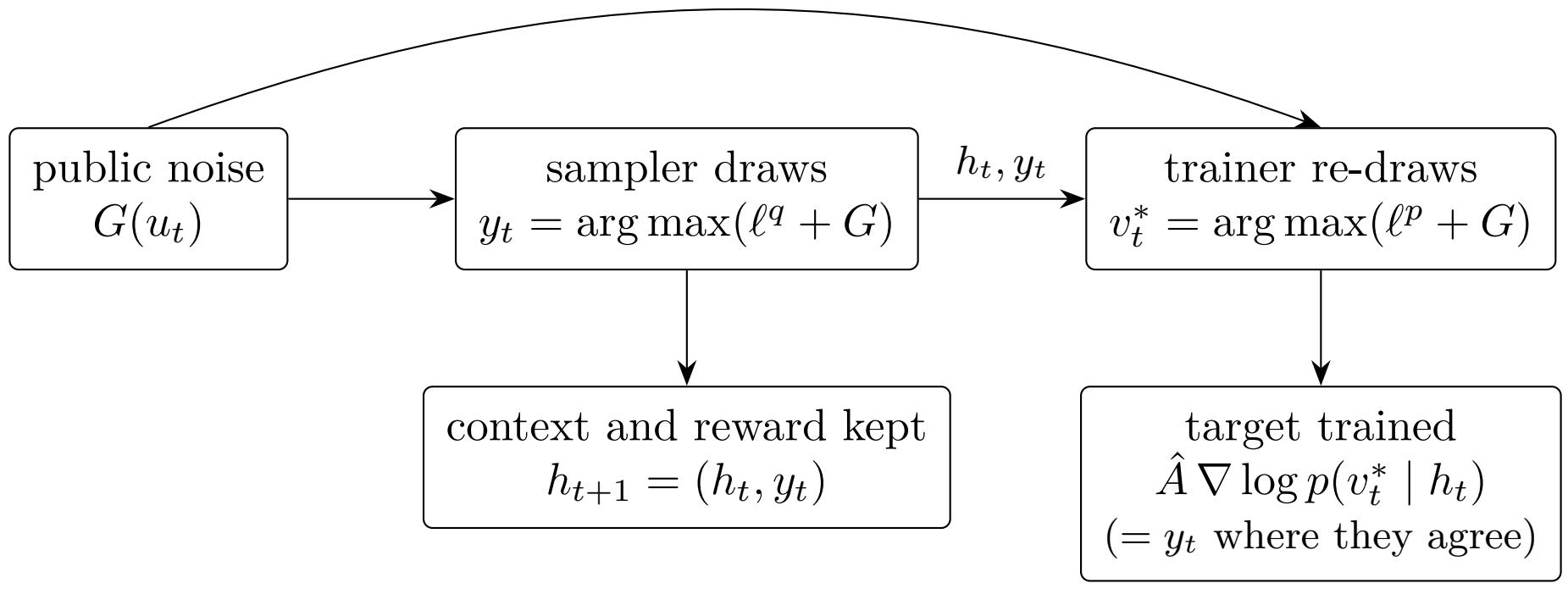
\begin{figure}[H]\centering
\begin{tikzpicture}[font=\small, node distance=9mm and 13mm,
  box/.style={draw, rounded corners=2pt, align=center, minimum height=10mm, inner sep=5pt},
  arr/.style={-{Stealth[length=2mm]}}]
\node[box] (noise) {public noise\\$G(u_t)$};
\node[box, right=of noise] (sampler) {sampler draws\\$y_t=\argmax(\ell^q+G)$};
\node[box, right=of sampler] (trainer) {trainer re-draws\\$\vstar_t=\argmax(\ell^p+G)$};
\node[box, below=of sampler] (ctx) {context and reward kept\\$h_{t+1}=(h_t, y_t)$};
\node[box, below=of trainer] (tgt) {target trained\\$\hat A\,\nabla\log p(\vstar_t\mid h_t)$\\{\footnotesize ($=y_t$ where they agree)}};
\draw[arr] (noise) -- (sampler);
\draw[arr] (noise.north) to[bend left=20] (trainer.north);
\draw[arr] (sampler) -- node[above, font=\footnotesize]{$h_t, y_t$} (trainer);
\draw[arr] (sampler) -- (ctx);
\draw[arr] (trainer) -- (tgt);
\end{tikzpicture}
\caption{Gumbel substitution at one position. The sampler's tokens and reward are kept as context; the target is always the trainer's own pick under the same noise.}
\label{fig:mechanism}
\end{figure}

\subsection{Public Gumbel noise}\label{sec:noise}
\begin{enumerate}
\item \textbf{Committed secret per checkpoint.} When checkpoint $k$ is adopted, the coordinator draws a 32-byte secret $S_k$ and publishes the commitment $H(S_k, k)$ immediately. $S_k$ is revealed once $k$ is more than 16 checkpoints old; a revealed checkpoint can no longer be leased.
\item \textbf{Lease nonce.} $\text{nonce} = \mathrm{SHA256}(\text{tag} \,\|\, S_k \,\|\, \text{env} \,\|\, \text{prompt} \,\|\, k)$. It does not depend on the sampler: two leases on the same (env, prompt, $k$) share the same draw.
\item \textbf{Uniform per rollout and position.} $u_t = \mathrm{SHA256}(\text{tag} \,\|\, \text{nonce} \,\|\, \text{prompt} \,\|\, k \,\|\, \text{rollout} \,\|\, t)[{:}8] / 2^{64}$, where $t$ counts the model's own tokens. In multi-turn episodes, tool observations and markup occupy no position.
\item \textbf{Gumbel per token.} splitmix64 mixes $\mathrm{key}(u_t)$ with the token id $v$ into a 53-bit uniform $U_v$, and $G_v = -\log(-\log U_v)$ is computed in float32 with a non-fused \texttt{logf}. The resulting bits are identical on A100, RTX 4090, H100 PCIe and CPU. A fused Triton kernel produces the same bits in 0.12\,ms for 64 rows and 0.34\,ms for 256 rows on an RTX 4090, against 10.9 and 43.7\,ms for the PyTorch reference.
\item \textbf{Pick and gap.} With $s = \ell + G$ in nats,
\begin{equation}
\vstar = \argmax_{v}\ \big(\ell_v + G_v\big), \qquad \mathrm{gap}(y) = \max_v s_v - s_y .
\end{equation}
Ties go to the lowest index; the gap is 0 when $y$ is the pick. This is the Gumbel-max trick~\cite{gumbel,maddison}: with independent standard Gumbel noise, $\Pr[\vstar = v] = \mathrm{softmax}(\ell)_v$, so the pick is an exact sample from the model's distribution. Once $G$ is fixed, the pick is a deterministic function of the logits, and a perturbation of one logit moves only that token's score: two parties with the same $G$ and logits $\ell^p$, $\ell^q$ can pick different tokens only if the two best scores are within $2\max_v|\ell^p_v - \ell^q_v|$ of each other.
\item \textbf{Sampling settings.} Temperature 1, no top-$p$ or top-$k$ truncation, set by the reference sampler, so the pick is an exact sample from the full-vocabulary softmax. The trainer picks $\vstar$ with an fp32 output head (an fp32 head matmul, with TF32 on bf16 operands). The reference sampler, built on vLLM~\cite{vllm}, also computes its head in fp32; a sampler may run a bf16 head or another precision instead, which lowers its match rate (Section~\ref{sec:hw}).
\item \textbf{Position indexing on the sampler.} The sampler wraps vLLM's model runner and indexes positions by request from the runner's token counts. A first version counted sampler calls per slot; vLLM preemption reset that counter, which misaligned noise and penalized small-memory cards (Appendix~\ref{app:eng}). Speculative decoding is disabled and the engine version is pinned.
\end{enumerate}

\subsection{Substitution loss}\label{sec:loss}
Let $h_{i,t}$ be the sampler's prefix at position $t$ of rollout $i$, $y_{i,t}$ the sampler's token and $\theta$ the latest checkpoint. The target is computed without gradient, and the loss scores it in the sampler's context:
\begin{equation}
\begin{aligned}
\vstar_{i,t} &= \argmax_v \Big[\ell^{\mathrm{fp32}}_{\theta}(h_{i,t})_v + G_v(u_{i,t})\Big],\\
L(\theta) &= -\frac{1}{N}\sum_{g}\sum_{i \in g} \hat{A}_i \sum_t \log p_\theta\big(\vstar_{i,t} \mid h_{i,t}\big).
\end{aligned}
\end{equation}
\begin{itemize}
\item At matching positions ($\vstar = y$), this is the usual on-policy score.
\item At disagreement positions, the trainer scores its own pick in the sampler's context, with the advantage earned by the sampler's trajectory.
\item \textbf{Advantage.} $\hat{A}_i = (\tilde r_i - \mathrm{mean}\,\tilde r)/\mathrm{std}(\tilde r)$ over a group of 16 rollouts, where $\tilde r_i$ is the reward after an overlong penalty ramping to 0.1 over the last 4,096 tokens~\cite{dapo}. Groups whose reward standard deviation is below 0.24 are not trained; with binary rewards this means groups where all 16 rewards are equal (one success in 16 already gives 0.242).
\item \textbf{Normalization.} Token mean over the whole step (512 groups), with $N$ fixed before the forward pass.
\item \textbf{Ratio and KL.} $\pi_{\text{old}}$ is the detached current policy and a checkpoint is published after every optimizer step, so the importance ratio is exactly 1 and PPO clipping is inactive. No KL penalty ($\beta=0$).
\end{itemize}

\subsection{Match rate, gates and payment}\label{sec:gates}
The match rate of a rollout is the fraction of stochastic positions ($\max p < 0.99$) where $\vstar = y$. It is computed during the training forward pass at almost no extra cost: one argmax per position over logits the forward already produces. It also serves as the verification of the rollout: unlike TOPLOC~\cite{toploc}, which commits to activations and needs its own proof, it checks the sampled tokens themselves, and the training rollouts carry no separate proof. Audited groups, which are not trained, need a separate forward.
\begin{itemize}
\item \textbf{Admission.} A rollout enters the loss if its match rate is at least 0.90, its maximum gap at most 8 nats, and its final end-of-sequence gap at most 1. A group is rejected if its pooled match rate falls below 0.90 or any rollout falls below 0.5.
\item \textbf{Suspicious positions.} Positions with gap above 3, or where $\vstar$ is a stop token that the sampler did not emit, trigger re-grading of the reward with the trainer's picks. Re-grading runs after the optimizer step, so it changes payment, not training.
\item \textbf{Payment.} Matched tokens of admitted rollouts $\times$ a per-environment price. A substituted position earns nothing, so lag and precision are paid for in proportion to the disagreement they cause.
\item \textbf{Audits.} Groups whose reward standard deviation is below 0.24 carry little or no GRPO signal; they are not trained and 15\% of them are audited against the latest checkpoint.
\end{itemize}
All thresholds are provisional, calibrated at learning rate $2\times10^{-6}$ (Section~\ref{sec:gapcal}). The full gate list is in Appendix~\ref{app:gates}.

\section{Analysis}\label{sec:analysis}

\emph{The drift term is zero in expectation for any sampler; the learning signal is not unbiased, and we say where the bias sits.}

\subsection{What is guaranteed}\label{sec:guaranteed}
\begin{proposition}[Zero drift]
The uniform $u_{i,t}$ enters only the hash of position $t$. For a sampler that draws each token with the noise of its own position, $G(u_{i,t})$ is independent of the prefix $h_{i,t}$, so Gumbel-max gives an exact sample: $\vstar \mid h \sim p_\theta(\cdot \mid h)$. For any coefficient $c$ independent of $u_{i,t}$ given the prefix (for example a function of the prefix, or of other rollouts' rewards),
\begin{equation}
\mathbb{E}_{u}\big[c\,\nabla \log p_\theta(\vstar \mid h)\big] \;=\; c \sum_v p_v\,\nabla \log p_v \;=\; 0 .
\end{equation}
\end{proposition}
The constant (baseline) part of the advantage therefore contributes nothing in expectation, whatever $q$ is and however $y$ and $\vstar$ are coupled. This holds for any hardware, precision and lag. The SFT-toward-$q$ term of Section~\ref{sec:problem} is removed without using $q$.

\paragraph{Scope: admission and the group baseline.} The identity holds before admission. The gates of Section~\ref{sec:gates} keep or drop a rollout according to its match rate, gaps and final token, all of which depend on the noise. Let $M\in\{0,1\}$ be the admission of a rollout and $g=\sum_t \nabla \log p_\theta(\vstar_t \mid h_t)$ its constant-coefficient term. Since $\mathbb{E}[g]=0$,
\begin{equation}
\mathbb{E}[M g] \;=\; -\,\mathbb{E}[(1-M)\,g], \qquad \big\|\mathbb{E}[M g]\big\| \;\le\; \Pr(M=0)\;\mathbb{E}\big[\|g\| \,\big|\, M=0\big].
\end{equation}
The residual is bounded by the excluded fraction times the gradient size of excluded rollouts. Excluded rollouts are those with more disagreements, so we expect the residual to tilt the retained targets toward the sampler's tokens, that is, toward $q$. Over the 131 production steps with full telemetry, the match, gap and end-of-sequence gates excluded 1.46\% of forwarded rollouts, and group rejection at most another 1.0\% (a rejected group is cancelled by a second pass with negated advantages, which is equivalent to excluding it). We therefore do not claim exact zero drift for the filtered estimator, and we have not measured this residual directly. Separately, the GRPO baseline is the group mean, which includes the rollout's own reward: that $1/16$ share, and the group standard deviation, which also includes it, depend on the rollout's own noise and belong with the cross-credit term below. Dropping groups with all-equal rewards adds nothing to this term, since their coefficient is zero; with non-binary rewards (overlong penalty, partial scores), the 0.24 threshold also drops some groups with nonzero advantages, a reward-dependent selection of the same kind as admission.

\subsection{What is not guaranteed}\label{sec:notguaranteed}
\begin{enumerate}
\item \textbf{Not an unbiased on-policy gradient.} Prefixes (the state distribution) and returns come from $q$, with no importance correction.
\item \textbf{Cross-credit bias.} At a disagreement position, $\hat{A}_i$ depends on $y$ and on the continuation, both coupled to $u_t$. With $\pi(v,y)$ the joint distribution of $(\vstar, y)$ and $Q$ the action value, the residual bias is
\begin{equation}
\sum_{y}\sum_{v \neq y} \pi(v, y)\,\nabla \log p_v\,\big[Q(y) - Q(v)\big].
\end{equation}
It is concentrated on disagreement positions and vanishes where $Q(y)=Q(v)$. Section~\ref{sec:stale} measures $|Q(\vstar)-Q(y)|$ directly.
\item \textbf{Different heads.} $\vstar$ is picked with the fp32 head, while the log-probability carrying the gradient comes from the model's bf16 head. The identity is exact for the fp32 softmax $p^{32}$ and approximate for the bf16 one $p^{16}$: the expected constant-advantage update on the logits is $p^{32} - p^{16}$ per position, a pull of the bf16 head toward the fp32 softmax of the same weights rather than toward the sampler. On 8.6M production positions replayed at checkpoints 156--157 (22 samplers, 7 environments), the fp32 and bf16 picks differ at 0.56\% of positions (1.08\% of stochastic positions) and the mean $\|p^{32}-p^{16}\|_1$ is 0.0098 per position (0.019 on stochastic positions), from 0.0037 (multi-turn tool use) to 0.015 (hard math and code) across environments. The deployed claim is therefore approximate zero drift, with a residual $L_1$ gap of about 0.01 per position, that is, about 0.5\% of probability mass moved.
\item \textbf{Reweighting.} Where picks agree, tokens are weighted by $p$ rather than $q$, as in on-policy training.
\end{enumerate}

\subsection{Threat model}\label{sec:threat}
The trainer and the coordinator's secrets are trusted. Nothing the sampler reports is: not its probabilities, its hardware, its precision or the checkpoint it used. Proposition~1 assumes causal generation, that is, each prefix is built without access to the noise of the position it precedes. Table~\ref{tab:threat} lists what a sampler can do and what the protocol guarantees in each case.

\begin{table}[H]\centering\small
\begin{tabular}{>{\raggedright\arraybackslash}p{0.38\linewidth}>{\raggedright\arraybackslash}p{0.55\linewidth}}
\toprule
Sampler behavior & Status \\
\midrule
Chooses its hardware, precision or an older checkpoint & Covered by Proposition~1, for any $q$. Paid only for matched tokens. \\\addlinespace
Reports false probabilities or metadata & No effect: the protocol reads tokens only. \\\addlinespace
Emits tokens that do not follow the draw & Each such token is substituted and unpaid; rollouts below match 0.90 or above gap 8 are excluded. No general detection guarantee for a forger who stays within the gates. \\\addlinespace
Chooses its prefix using the noise of later positions & Violates the independence assumption. Possible in principle, since the lease nonce determines the noise of the whole rollout. Deviations are limited by the gates and each one is substituted and unpaid, but they are not excluded. \\\addlinespace
Withholds a group after seeing its draws & The noise is fixed per prompt, checkpoint, rollout index and position, so a sampler cannot redraw; it can only forfeit a whole leased group and its pay. This is a selection the bound above does not cover. \\
\bottomrule
\end{tabular}
\caption{Threat model.}
\label{tab:threat}
\end{table}
A stronger guarantee would reveal each position's noise only after the prefix is committed, which conflicts with fully asynchronous generation. Public-noise verification and the gates are operational checks; they do not establish unbiasedness against adaptive samplers.

\subsection{Coupling quality and the meaning of $\vstar$}\label{sec:coupling}
At a single position, the sampler and the trainer sample from $q$ and $p$ with shared noise and no communication. No coupling agrees more often than $1-\mathrm{TV}(p,q)$; Gumbel sampling agrees at least $(1-\mathrm{TV})/(1+\mathrm{TV})$ of the time~\cite{daliri}:
\begin{equation}
\mathrm{TV}(p, q) \;\le\; \Pr[\vstar \neq y] \;\le\; \frac{2\,\mathrm{TV}(p, q)}{1 + \mathrm{TV}(p, q)} .
\end{equation}
The disagreement rate therefore bounds the sampler--trainer mismatch within a factor of two, without any reported probability. Under the Gumbel-max structural causal model, $\vstar$ is the counterfactual token: the one $p$ would have emitted in the same random circumstances as $y$~\cite{oberst}. Substitution trains on that counterfactual without regenerating the continuation.

\section{Controlled experiments}\label{sec:lab}

\emph{In the lab, substitution removes the measured drift, keeps most of the gradient, and prevents the collapses seen without correction.}

\paragraph{Setup.} Losses: L0, no correction (score $y$); L1, mask (train only where $y=\vstar$); L2, substitution; L3, MIS $[0.2, 5]$ + score centering using the true $q$~\cite{liu,centering}. Models: ``9B'' is \texttt{teutonic-i-attention-\allowbreak graft-healed-v1}, a base model without SFT unless stated; ``2B'' is Qwen3.5-2B. Match rates are measured on stochastic positions.

\subsection{Gumbel-max against inverse-CDF sampling}\label{sec:gumbel-vs-icdf}
\emph{Gumbel-max disagrees 2.1 to $2.6\times$ less often than inverse-CDF sampling in token-id order, and stays within the worst-case guarantee of communication-free coupling.}

\begin{table}[H]\centering\small
\resizebox{\linewidth}{!}{%
\begin{tabular}{lccc}
\toprule
Condition & TV$(p,q)$ & Disagreement, inverse-CDF & Disagreement, Gumbel-max \\
\midrule
Honest, bf16 heads & 0.0122 & 0.0373 & 0.0155 \\
Honest, fp32 heads & 0.0044 & 0.0218 & 0.0085 \\
fp8 weights & 0.0318 & 0.0806 & 0.0388 \\
\bottomrule
\end{tabular}
}
\caption{Per-token disagreement rate between sampler and trainer under the two sampling schemes, on stochastic positions.}
\end{table}
Inverse-CDF sampling draws one uniform $u_t$ and returns the first token, in token-id order, whose cumulative probability exceeds $u_t$. The interval of token $v$ starts at $\sum_{w<v} p_w$, so small differences between $p$ and $q$ on every lower-index token, most of them unlikely, accumulate and shift the boundaries of likely tokens. Inverse-CDF disagreements are therefore largely an artifact of token-id order. Under Gumbel-max a difference on one logit moves only that token's score, and the disagreement rate of a token depends only on its probability $p_y$: across 881 tokens, none deviates from the rate predicted by $p_y$ by more than four standard deviations. This is why the deployed sampler uses Gumbel-max. The measured Gumbel-max rates sit between the two bounds of Section~\ref{sec:coupling}, at 1.2 to $1.9\times$ TV$(p,q)$.

\subsection{Match rate across hardware, precision and environments}\label{sec:hw}
\emph{Honest samplers on five card types agree to within 0.0006; 4-bit quantization falls below the 0.90 admission floor.}

\begin{table}[H]\centering\small
\begin{tabular}{lc}
\toprule
Sampler $\to$ verifier & Match rate \\
\midrule
RTX 3090, 5090, A6000, L40, H100; bf16 heads & 0.9858--0.9864 \\
fp32 heads on both sides & 0.9939--0.9942 \\
fp8 KV cache & 0.9851--0.9856 \\
W8A16 (Marlin) / online fp8 weights & 0.970 / 0.963--0.970 \\
W4A16 / NVFP4 & 0.854--0.856 \\
\bottomrule
\end{tabular}
\caption{Match rate of honest and quantized samplers against the trainer.}
\end{table}
\begin{itemize}
\item \textbf{Environments.} DAPO math 0.9866, hard math 0.9863, code 0.9854, instruction following (IF) 0.9838. Under inverse-CDF, IF trailed math and code by about 9 points and an earlier gate rejected 55 of 64 honest IF groups; under Gumbel the gap is about 0.2 points.
\item \textbf{Detectability.} fp8 weights are separated from honest rollouts with per-rollout AUC 0.98--0.997 under Gumbel (0.82--0.86 under inverse-CDF). An fp8 KV cache is indistinguishable (AUC 0.55), and its drift is removed like any other.
\item \textbf{Payment floor.} Paying per matched token without a floor would favor quantization: W4A16 on an RTX 3090 earns 59\% more paid tokens per second. fp8 weights, at 0.963--0.970, stay above the floor by design: their drift is removed like any other, and they lose the pay of every token they disagree on. A per-rollout floor of 0.95 excludes all W4A16 and keeps every honest lag-0 rollout. The deployed floor is 0.90, lowered to tolerate lag (Appendix~\ref{sec:qualif}); W4A16, at about 0.855, stays below it.
\item \textbf{Ceiling.} vLLM against itself on the same card reaches about 0.996: decoding is not batch-invariant, and only 67 of 128 regenerations of the same rollout are identical.
\item \textbf{Cost to the sampler.} The forced Gumbel draw costs 0.7\% of throughput in bf16 and 1.5--1.7\% in fp8 on an RTX 4090.
\end{itemize}

\subsection{Drift at a fixed checkpoint}\label{sec:drift}
\emph{Without correction, the expected update is exactly SFT toward the sampler; with substitution it stays at or near the numerical floor in every condition.} The expected drift is the norm of the constant-advantage gradient summed over the measured positions, in arbitrary but common units; the floor is the value it keeps, about 100, when nothing is left to remove, so the ratios to L0 are lower bounds. These measurements use the same head on both sides; the deployed fp32/bf16 head gap adds a residual of its own (Section~\ref{sec:notguaranteed}). The measurements in this section and the gradient cosines of Section~\ref{sec:stale} were made with inverse-CDF sampling, before the switch to Gumbel-max; the zero-drift identity holds for any coupling, but the signal figures are not re-measured under Gumbel.

\begin{table}[H]\centering\small
\resizebox{\linewidth}{!}{%
\begin{tabular}{lcccc}
\toprule
Condition & L0 bias $\|b\|_1$ & L2 bias $\|b\|_1$ & L0 expected drift & L2 expected drift \\
 & per position & per position & norm (cos.\ with SFT$\to q$) & norm \\
\midrule
Honest, bf16 & 0.0152 & 0.0001 & 16,382 (1.000) & 96.5 \\
Honest, fp32 & 0.0057 & 0.0001 & 17,171 (1.000) & 93.8 \\
fp8 weights & 0.0406 & 0.0001 & 111,861 (1.000) & 97.7 \\
Weight noise $\sigma=0.06$ & 0.1758 & 0.0002 & 640,223 (1.000) & 133.1 \\
\bottomrule
\end{tabular}
}
\caption{Expected drift at a fixed checkpoint (9B, inverse-CDF sampling). $\|b\|_1$ is averaged over all positions, so it equals $2\,\mathrm{TV}$ there, not on stochastic positions as in the previous table.}
\end{table}
\begin{itemize}
\item \textbf{Masking is worse than nothing.} On a separate honest-sampler measurement, L1 increases the constant-advantage gradient norm to 244k against 69k for L0.
\item \textbf{Signal kept.} For an honest sampler, the cosine between the L2 and L0 gradients is 0.938. In that measurement L3 coincides with L0, so the figure says how much of the uncorrected direction substitution keeps when there is little drift to remove; it does not measure closeness to the true-$q$ correction. Disagreement positions carry 5.5\% of the gradient ($|A|\cdot\|e_y-p\|$), each about $4\times$ the mean per-position norm. The position-mass proxy (1 $-$ weighted share of disagreements) is 97.9\% overall, 96.5\% on stochastic positions and 93.8\% entropy-weighted in bf16; it is a proxy, not a gradient measurement.
\end{itemize}

\subsection{Staleness and the cross-credit bias}\label{sec:stale}
\emph{Lag lowers the match rate and rotates the gradient away from the true-$q$ correction, while substitution's drift stays at the floor; disagreement positions are not more valuable than matched controls.}

The 9B model was trained for 16 steps at learning rate $2\times10^{-6}$; rollouts from earlier checkpoints are scored against the latest one. The Gumbel column replays the same rollouts under Gumbel noise; the replay matches real Gumbel runs (0.98625 against 0.98603).

\begin{table}[H]\centering\small
\setlength{\tabcolsep}{4pt}
\resizebox{\linewidth}{!}{%
\begin{tabular}{cccccccc}
\toprule
Lag & Match, & Match, Gumbel & Gradient & L0 expected & L2 expected & $\cos$(L2, L3 & $\cos$(L0, L3 \\
(steps) & inverse-CDF & (replay) & rewritten & drift & drift & true $q$) & true $q$) \\
\midrule
0 & 0.973 & 0.987 & 5.5\% & -- & -- & 0.938 & 1.00 \\
1 & 0.962 & 0.983 & 7.7\% & 59k & floor & 0.899 & 0.981 \\
2 & 0.961 & 0.982 & 8.0\% & 87k & floor & 0.908 & 0.98 \\
4 & 0.945 & 0.970 & 11.5\% & 418k & floor & 0.866 & 0.752 \\
8 & 0.912 & 0.949 & 17.9\% & 900k & floor & 0.792 & 0.606 \\
\bottomrule
\end{tabular}
}
\caption{Effect of checkpoint lag (9B).}
\end{table}
Gumbel reduces the match loss per step of lag by about 40\% without removing it. On this run, L0 is closer to the true-$q$ gradient up to lag 2 and L2 from lag 4. On a second run (checkpoint 8 to 16), L0 stays closer even at lag 8 because its drift is $6\times$ smaller. The crossover depends on the path the policy has taken, not on the number of steps.

\paragraph{Branching test (9B, math).} From each branch point we run 8 to 16 continuations and estimate $|Q(\vstar)-Q(y)|$. With so few continuations, estimation noise inflates the absolute values; the comparison with matched controls, which carry the same noise, is what the test measures.
\begin{table}[H]\centering\small
\begin{tabular}{lcc}
\toprule
Positions & $n$ & Mean $|Q(\vstar)-Q(y)|$ [95\% CI] \\
\midrule
Disagreement positions & 1,000 & 0.096 [0.089, 0.103] \\
Entropy-matched controls & 1,000 & 0.103 [0.096, 0.111] \\
\bottomrule
\end{tabular}
\caption{Value difference at disagreement positions and at matched controls.}
\end{table}
Disagreements are not concentrated on decision points, so the cross-credit bias of Section~\ref{sec:notguaranteed} is not amplified there. This test covers math only.

\subsection{Calibrating the gap test}\label{sec:gapcal}
On a 9B SFT model with fp32 heads, honest lag-0 rollouts on an RTX 4090 have a p99 gap of 0.127 nats and a maximum of 0.191; fp8 weights reach 2.18. Lag is simulated by $n$ sign-of-Adam steps on mixed gsm8k, code and IF gradients:
\begin{table}[H]\centering\small
\resizebox{\linewidth}{!}{%
\begin{tabular}{lccc}
\toprule
Environment & Max gap at lag 1 / 2 / 4 (LR $2\times10^{-6}$) & Rejected at gap 8 & Re-graded at gap 3 (reward changed) \\
\midrule
gsm8k & 0.19 / 0.40 / 0.82 & 0 & 0 \\
Code & 0.36 / 0.68 / 3.86 & 0 & 2 of 384 (0) \\
IF & 2.06 / 1.90 / 2.21 & 0 & 0 \\
\bottomrule
\end{tabular}
}
\caption{Gap under simulated lag.}
\end{table}
At lag 4 and learning rate $5\times10^{-6}$ on code, 4 of 384 rollouts are rejected and 71 re-graded: the 8 / 3 / 1 thresholds are safe at $2\times10^{-6}$, not at $5\times10^{-6}$. For the final end-of-sequence token, a probability floor would reject honest IF rollouts ($p_{\text{stop}}$ down to $6.6\times10^{-7}$); a gap rule does not.

\subsection{Mini-trainings}\label{sec:minitrain}
\emph{Uncorrected training looks better at first, then collapses; substitution stays stable and is comparable to the true-$q$ correction, but costs reward on a short horizon with a realistic fp8 sampler.}

2B, gsm8k, Gumbel noise, 120 steps unless stated. ``Strong'' mismatch: weight noise calibrated to 0.90 match rate plus lag 4. ``Real fp8'': vLLM fp8 plus lag 4. Two values in a cell are two seeds.

\begin{figure}[H]\centering
\includegraphics[width=\linewidth]{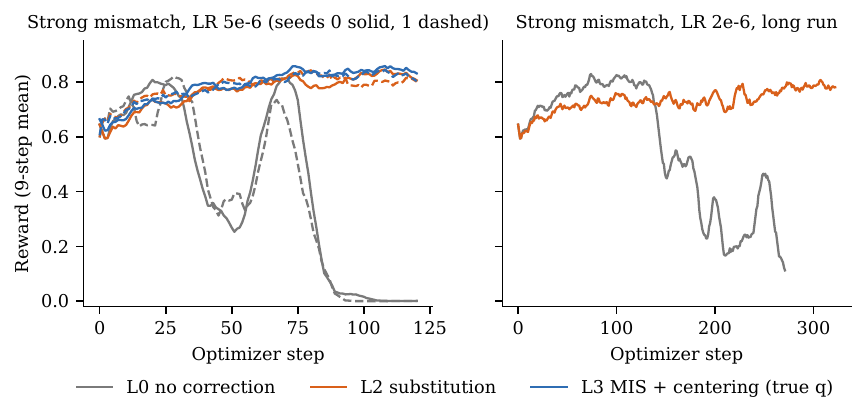}
\caption{Reward of the mini-trainings under strong mismatch (2B, gsm8k). Left: learning rate $5\times10^{-6}$, two seeds; uncorrected training (L0) collapses at steps 41 and 44. Right: long run at $2\times10^{-6}$; L0 collapses at step 184, substitution (L2) stays stable for 323 steps.}
\label{fig:collapse}
\end{figure}

\begin{table}[H]\centering\small
\setlength{\tabcolsep}{4pt}
\resizebox{\linewidth}{!}{%
\begin{tabular}{lllcccl}
\toprule
Mismatch & Loss & LR & Reward, last 20 & Greedy eval & Entropy & Outcome \\
\midrule
None & L0 & 2e-6 & 0.778 & 0.798 & 0.351 & Reference (2 seeds) \\
Strong & L0 & 2e-6 & 0.791 & 0.815 & 0.221 & Best short-term, entropy collapsing \\
Strong & L2 & 2e-6 & 0.734 & 0.770 & 0.341 & \\
Strong & L3 & 2e-6 & 0.740 & 0.780 & 0.334 & \\
Strong & L0 & 5e-6 & 0.001 / 0.000 & 0.000 & 0.08 / 0.02 & Collapse at steps 41, 44 \\
Strong & L2 & 5e-6 & 0.824 / 0.805 & 0.815 / 0.840 & 0.30 / 0.24 & Stable \\
Strong & L3 & 5e-6 & 0.844 / 0.824 & 0.870 / 0.855 & 0.26 & Stable \\
Strong, long & L0 & 2e-6 & 0.184 (272 steps) & 0.245 & 0.055 & Collapse at step 184 \\
Strong, long & L2 & 2e-6 & 0.779 (323 steps) & 0.805 & 0.354 & Stable, KL 0.019--0.029 \\
Real fp8 & L0 & 2e-6 & 0.816 & 0.860 & 0.387 & No collapse in 120 steps \\
Real fp8 & L2 & 2e-6 & 0.765 & 0.795 & 0.339 & $-0.05$ reward vs.\ L0 \\
\bottomrule
\end{tabular}
}
\caption{Mini-trainings (2B, gsm8k). In the L0 collapses at $5\times10^{-6}$, KL rises to 1.5--3.0. Under real fp8 the match rate is about 0.94--0.96.}
\end{table}
\begin{enumerate}
\item Short-horizon reward is misleading: L0 leads while its entropy collapses, then diverges.
\item L2 performs comparably to L3 in these settings (one or two seeds), without any information about $q$; L3 leads by about 0.02 reward at $5\times10^{-6}$, seed for seed.
\item With a realistic fp8 sampler on a short horizon, substitution costs about 0.05 reward.
\end{enumerate}
The L3 arm of the real-fp8 round was stopped between steps 9 and 38 and is not reported. A run counts as collapsed at the first step where its 10-step moving-average reward falls below half the mean reward of steps 0--19; the smoothed reward of L0 at $5\times10^{-6}$ already falls below half from step 36. Figure~\ref{fig:collapse} uses the same 10-step smoothing. L3 at $5\times10^{-6}$ ran on H100, L0 and L2 on A100; the long runs stopped at 272 (L0) and 323 (L2) of 400 planned steps.

\section{Production run}\label{sec:prod}

\emph{Over 158 optimizer steps and 15.75B tokens on a permissionless network, no step was skipped or clipped and the model improved on all eight held-out benchmarks.}

Run \texttt{teutonic-rl-async-v1}, steps 1 to 158, from October 22 to November 2, 2026, on Reliquary. The starting model is Teutonic-I, our 9B model after supervised fine-tuning. Training mixed 7 environments: instruction following, hard math, competitive code, DAPO math, SWE tasks, TMax terminal tasks and EnvScaler multi-turn tool use. The trainer ran on a single 8-GPU node (H100 or H200; the node changed during the run). Data from W\&B (trainer and coordinator) and the evaluation reports. Final totals cover the full run; the per-phase telemetry in Section~\ref{sec:stability} covers steps 1--131. Averages are weighted by trained tokens per step unless stated.

\subsection{Scale}
\begin{table}[H]\centering\small
\begin{tabular}{lp{0.55\linewidth}}
\toprule
Quantity & Value \\
\midrule
Optimizer steps & 158 over 11 days \\
Trained tokens & 15.75B \\
Samplers & 161 accounts with at least one trained group; up to 73 active at once over steps 1--131 \\
Match rate / substitution & 0.979 (stochastic positions) / 1.21\% (all trained tokens) \\
Step duration (median, steps 1--131) & 68 min; 104 min over steps 1--20, 59 min over steps 61--131 as completions shortened (120M to 74M trained tokens per step) \\
Declared hardware (opt-in) & RTX 5090, RTX PRO 6000 Blackwell, B200, RTX 3090, H200, RTX 4090, RTX A6000, A100, RTX 6000 Ada; about 77\% of trained tokens over steps 1--131 come from undeclared samplers \\
\bottomrule
\end{tabular}
\caption{Scale of the production run.}
\end{table}

\paragraph{Where the time went.} No step waited for a slow or lagging sampler: rollouts were accepted up to 16 checkpoints behind, the mean lag per step was 2.3 checkpoints, and cards from RTX 3090 to B200 contributed to the same steps. A step starts as soon as 512 groups are buffered. Over steps 8--131, the trainer spent 87\% of its running time in optimizer steps, 2\% in audits run before each step, and 10\% waiting for the batch to fill, most of which it used for further audits. Step time follows the number of trained tokens per step, at 20,000 to 25,000 trained tokens per second. Because substitution keeps stale rollouts usable, the trainer never waits for a lagging sampler; what bounds the step time is trainer compute.

\subsection{Optimization stability}\label{sec:stability}
\begin{figure}[H]\centering
\includegraphics[width=\linewidth]{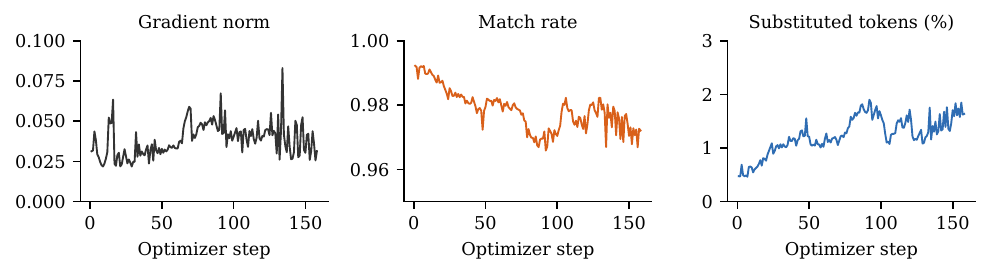}
\caption{Production telemetry per optimizer step: gradient norm, match rate, and share of trained tokens substituted.}
\label{fig:stability}
\end{figure}

No step was skipped for a gradient spike or non-finite values over the 158 steps. The maximum gradient norm is 0.083, 12$\times$ below the clipping threshold. There is no KL term and the PPO ratio is 1. Over the full run, entropy declines from 0.70 to 0.54--0.57 in the later phases (Figure~\ref{fig:stability}).

\begin{table}[H]\centering\small
\setlength{\tabcolsep}{3.5pt}
\resizebox{\linewidth}{!}{%
\begin{tabular}{lcccccccc}
\toprule
Steps & Match & Subst. & Entropy & Grad norm, median & Mean & $\|\Delta\theta\|/\|\theta\|$ & Re-graded & Below \\
 & rate & & (median) & [range] & lag & (median) & & floor \\
\midrule
1--20 & 0.9900 & 0.62\% & 0.695 & 0.030 [0.022--0.064] & 1.37 & 2.2e-5 & 1.3\% & 0.17\% \\
21--40 & 0.9831 & 1.03\% & 0.688 & 0.028 [0.022--0.043] & 1.52 & 7.7e-5 & 4.9\% & 2.7\% \\
41--60 & 0.9801 & 1.15\% & 0.634 & 0.033 [0.024--0.039] & 3.17 & 8.4e-5 & 5.6\% & 2.1\% \\
61--80 & 0.9779 & 1.24\% & 0.599 & 0.046 [0.033--0.059] & 1.40 & 1.3e-4 & 3.8\% & 1.0\% \\
81--100 & 0.9698 & 1.69\% & 0.541 & 0.046 [0.034--0.067] & 2.29 & 1.6e-4 & 6.2\% & 2.3\% \\
101--131 & 0.9774 & 1.32\% & 0.574 & 0.043 [0.030--0.055] & 3.02 & 1.1e-4 & 4.8\% & 0.86\% \\
All & 0.9808 & 1.12\% & & & & & 4.5\% & 1.5\% \\
\bottomrule
\end{tabular}
}
\caption{Telemetry by phase, steps 1--131.}
\end{table}
\begin{itemize}
\item \textbf{Gradient.} The norm stays between 0.022 and 0.067 over steps 1--131 with no spike. Its median rises by about half (0.030 to 0.043--0.046) while the relative update size grows 5 to $7\times$ ($2.2\times10^{-5}$ to $1.1$--$1.6\times10^{-4}$), as the learning-rate warm-up weighs on the first phase.
\item \textbf{Entropy} declines slowly from 0.70 to 0.54--0.57, with no collapse; part of the decline comes from the environment mix.
\item \textbf{Substitution} rises from 0.6\% to 1.0--1.7\% of trained tokens (phase averages). It is the only indicator that degrades clearly; the rewritten share stays below 2\% of tokens.
\end{itemize}

\subsection{Match rate over the run}
The match rate eases from 0.990 over the first 20 steps to 0.97--0.98 later in the run, mostly on agentic tasks and instruction following, while math and code stay near 0.98. Lag does not explain it: about 0.3 point between lag 0 and lag 3 or more over steps 16--131. Neither does position: 0.9800 below 1k tokens, 0.9791 beyond 32k. Substitution absorbs these disagreements, 1.2 to 1.9\% of trained tokens per step after step 20 (1.0 to 1.7\% averaged by phase), with no effect visible on gradient norm or entropy.

\subsection{Match-rate clusters across the sampler fleet}
\emph{Per-sampler match rates on math form tight clusters that fall on the values the lab measured for each precision (Section~\ref{sec:hw}), with an almost empty gap between 0.970 and 0.983.}

\begin{figure}[H]\centering
\includegraphics[width=\linewidth]{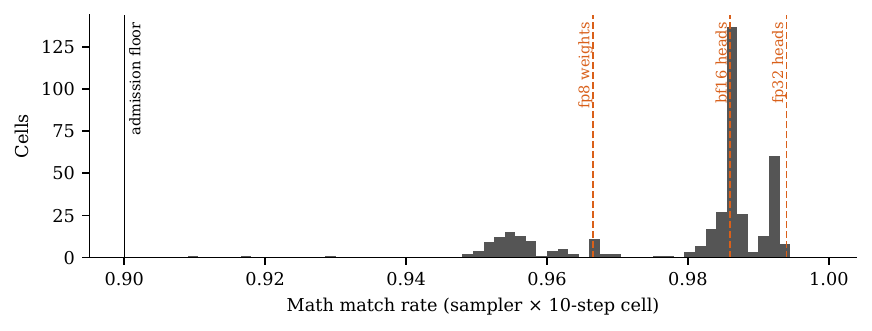}
\caption{Per-sampler math match rate over 407 sampler $\times$ 10-step cells. Dashed lines: lab values for each precision.}
\label{fig:samplers}
\end{figure}

\begin{table}[H]\centering\small
\resizebox{\linewidth}{!}{%
\begin{tabular}{lll}
\toprule
Production cluster (math match rate) & Declared hardware & Lab value for the matching precision \\
\midrule
$\sim$0.993 & Mostly B200 and H200 & fp32 heads: 0.994 \\
$\sim$0.986, the bulk & Almost all undeclared & bf16 heads: 0.986; fp8 KV cache: 0.985 \\
$\sim$0.967 & A few stable, undeclared samplers & fp8 weights: 0.963--0.970 \\
$\sim$0.955 & RTX 5090 & Unexplained: 0.01 below fp8 weights \\
\bottomrule
\end{tabular}}
\caption{Match-rate clusters and the lab values they fall on.}
\end{table}
Almost no sampler declares its precision. The clusters correspond to the lab's precisions; they do not identify any individual sampler's precision. During steps 3--6, a broken or forging sampler, with per-rollout match rates down to about 0.58, was stopped by the gates, which rejected 8 groups.

\subsection{Benchmarks}\label{sec:bench}
\emph{The RL run improved the model on all eight held-out benchmarks, most on agentic and code tasks.}

\paragraph{Protocol.} Every evaluation sample was generated by the subnet's samplers under an imposed Gumbel draw, verified against that draw, and refused, never corrected, if it did not follow it. A truncated or missing sample counts as a failure. All three models are sampled at temperature 1.0, top-$p$ 0.95, top-$k$ 20. The draw is then the Gumbel-max over the tokens kept after truncation, and the verifier recomputes it with the same truncation. Budgets: 64k tokens for AIME 2025, AIME 2026, GPQA Diamond~\cite{gpqa} and LiveCodeBench~\cite{lcb}; 32k for IFBench~\cite{ifbench} and MMLU-Pro~\cite{mmlupro}; 48 turns for SWE-bench Verified~\cite{swebench} (mini-swe-agent harness~\cite{sweagent}) and TMax. SWE-bench Verified uses 486 of its 500 problems: 7 need network access and 7 have a gold patch that fails in our sandbox. TMax is built from the terminal tasks of TMax-15K~\cite{tmax}, which we converted to run offline in our sandboxes and validated (4,998 of 14,601 tasks kept: the reference solution passes, an empty attempt fails, and mutated answers fail); we evaluate on 200 tasks of its held-out split \texttt{tmax\_sft} (2,386 tasks), which neither SFT nor RL used. Samples per problem: 8 for AIME, 4 for GPQA Diamond and TMax, 2 for the other sets. The reference model, Qwen3.5-9B, uses the same chat template, harness and protocol; it is not a controlled comparison, since its base model and post-training differ from ours.

\begin{table}[H]\centering\small
\resizebox{\linewidth}{!}{%
\begin{tabular}{lrcccc}
\toprule
Benchmark & $n$ & Teutonic-I, after SFT & Teutonic-I-Reliquary, after RL & $\Delta$ & Qwen3.5-9B \\
\midrule
SWE-bench Verified & 486 & 17.3 {\scriptsize[13.9--20.7]} & \textbf{43.6} {\scriptsize[39.2--48.0]} & +26.3 & 35.1 {\scriptsize[30.9--39.3]} \\
TMax & 200 & 28.4 {\scriptsize[22.2--34.6]} & 74.8 {\scriptsize[68.8--80.8]} & +46.4 & \textbf{82.2} {\scriptsize[76.9--87.5]} \\
LiveCodeBench & 300 & 40.7 {\scriptsize[35.1--46.3]} & \textbf{62.0} {\scriptsize[56.5--67.5]} & +21.3 & 57.0 {\scriptsize[51.4--62.6]} \\
AIME 2025 & 30 & 73.3 {\scriptsize[57.5--89.1]} & \textbf{91.2} {\scriptsize[81.1--100.0]} & +17.9 & 86.2 {\scriptsize[73.9--98.5]} \\
AIME 2026 & 30 & 74.2 {\scriptsize[58.5--89.9]} & \textbf{89.6} {\scriptsize[78.7--100.0]} & +15.4 & 86.7 {\scriptsize[74.5--98.9]} \\
IFBench & 300 & 65.3 {\scriptsize[59.9--70.7]} & \textbf{74.7} {\scriptsize[69.8--79.6]} & +9.4 & 67.5 {\scriptsize[62.2--72.8]} \\
MMLU-Pro & 500 & 74.5 {\scriptsize[70.7--78.3]} & 79.1 {\scriptsize[75.5--82.7]} & +4.6 & \textbf{80.3} {\scriptsize[76.8--83.8]} \\
GPQA Diamond & 198 & 64.8 {\scriptsize[58.1--71.5]} & 68.1 {\scriptsize[61.6--74.6]} & +3.3 & \textbf{81.6} {\scriptsize[76.2--87.0]} \\
\bottomrule
\end{tabular}
}
\caption{Final evaluation, pass@1 (\%). $n$ is the number of problems. Brackets: conservative 95\% intervals, $\hat p \pm 1.96\sqrt{\hat p(1-\hat p)/n}$, which upper-bound the variance of a mean of per-problem pass rates; they are wider than a bootstrap over problems.}
\label{tab:final}
\end{table}

\begin{figure}[H]\centering
\includegraphics[width=\linewidth]{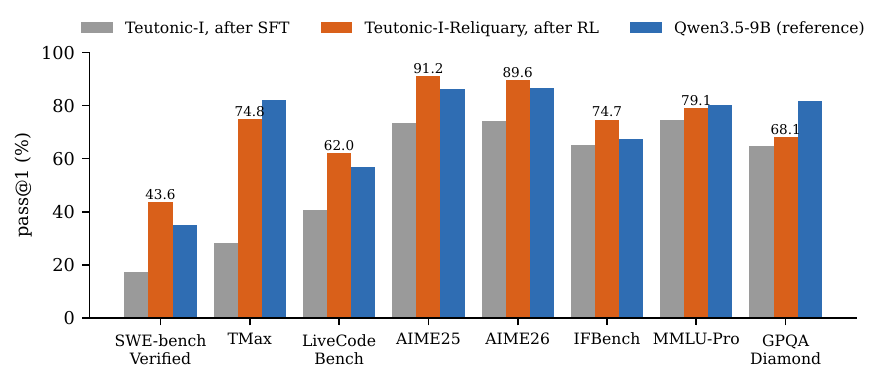}
\caption{Final evaluation, pass@1 (\%), same protocol for the three models.}
\label{fig:final}
\end{figure}

\begin{itemize}
\item \textbf{Gains.} RL improves every benchmark, from +3.3 points (GPQA Diamond) to +46.4 (TMax). The largest gains are on the agentic and code sets: SWE-bench Verified $\times2.5$, TMax $\times2.6$, LiveCodeBench +21.3, all three with disjoint intervals even under the conservative bound. On AIME (30 problems), IFBench, MMLU-Pro and GPQA the intervals overlap. SWE-bench Verified pass@2 reaches 51.0.
\item \textbf{Reference.} After RL the model is ahead of Qwen3.5-9B on 5 of 8 benchmarks (SWE-bench Verified, LiveCodeBench, AIME 2025 and 2026, IFBench) and behind on TMax, MMLU-Pro and GPQA Diamond. After SFT it was behind on all 8. Under the conservative bound only the GPQA Diamond gap is outside the intervals; on SWE-bench Verified (+8.5 points) the intervals meet within 0.1 point.
\item \textbf{Family overlap.} TMax and SWE-bench Verified are held out, but from the same task families as training (\texttt{tmax\_v1}, training repositories). Their gains partly measure learning of the family; the sets farthest from training (GPQA, MMLU-Pro) move least.
\item \textbf{Integrity.} Every sample of every report is verified against the draw. Samples left unsettled count as failures: TMax 1 and SWE-bench Verified 3 after RL, TMax 4 and SWE-bench Verified 2 after SFT.
\end{itemize}

\begin{figure}[H]\centering
\includegraphics[width=\linewidth]{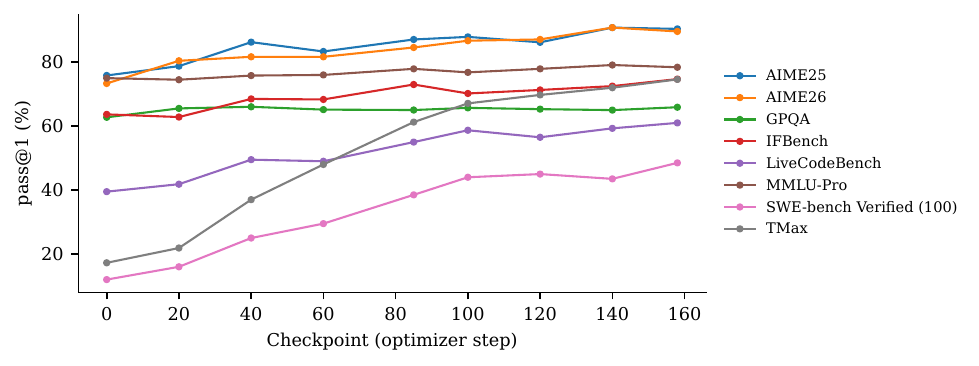}
\caption{pass@1 per checkpoint during the run, run-time protocol (not comparable to Table~\ref{tab:final}). Checkpoint 0 is the SFT model.}
\label{fig:ckpt}
\end{figure}

\paragraph{During the run.} While training, we evaluated checkpoints with an earlier version of the protocol: temperature 1, top-$p$ 1, top-$k$ 0, a 100-problem subset of SWE-bench Verified, and TOPLOC~\cite{toploc} in addition to the draw for checkpoints 0--140. These numbers are not comparable to Table~\ref{tab:final}; on the 99 problems the two SWE-bench sets share, checkpoint 158 scores 48.5 under the run-time protocol and 44.9 under the final one. They show the shape of training (Figure~\ref{fig:ckpt}, Appendix~\ref{app:ckpt}): SWE-bench Verified and TMax rise at nearly every checkpoint, and between checkpoints 0 and 85 the 95\% bootstrap intervals are disjoint for SWE-bench Verified, TMax, LiveCodeBench and IFBench. Under that protocol, pass@$k$ rises with pass@1 on TMax (pass@4 36.5 to 80.5 between checkpoints 0 and 85), SWE-bench Verified (pass@2 19.0 to 49.0), LiveCodeBench (49.7 to 61.7) and IFBench (77.0 to 86.7): RL widens what the model solves rather than only concentrating mass. RL also cuts truncated answers (AIME 18.8\% to 9.6\%, GPQA 8.3\% to 0.5\%, IFBench 7.7\% to 0.8\%) and shortens answers by 15 to 35\%; since truncation counts as failure, part of the gain on AIME, GPQA and LiveCodeBench comes from there.

\section{Related work}\label{sec:related}

\emph{Drift corrections need the sampler's probabilities, bitwise alignment needs identical kernels, and verification does not build a training target.}

\begin{table}[H]\centering\small
\begin{tabular}{p{0.22\linewidth}p{0.19\linewidth}p{0.12\linewidth}p{0.14\linewidth}p{0.16\linewidth}}
\toprule
Method & Removes drift & Needs $q(y)$ & Checks sampled tokens & Heterogeneous hardware \\
\midrule
TIS~\cite{yao}, MIS~\cite{liu}, IcePop~\cite{icepop} & Truncates or masks & Yes & No & Yes \\
GSPO~\cite{gspo}, CISPO~\cite{cispo} & Sequence-level or clipped weights & Yes & No & Yes \\
Score centering + MIS~\cite{centering,modal} & Yes, in expectation & Yes, top-$k$ of $q$ & No & Yes \\
Bitwise alignment~\cite{thinkingmachines,vllmbitwise} & Removes the mismatch itself & No & No & No: same kernels on both sides \\
TOPLOC, INTELLECT-2~\cite{toploc,intellect2} & No & No & No: checks activations & Within a tolerance \\
Token-DiFR~\cite{difr} & No: verification only & No & Yes & Yes \\
\textbf{Gumbel substitution} & \textbf{Yes, in expectation} & \textbf{No} & \textbf{Yes} & \textbf{Yes} \\
\bottomrule
\end{tabular}
\caption{How related approaches handle sampler drift on an open network.}
\end{table}

\paragraph{Mismatch corrections.} Yao et al.\ showed that inference and training engines turn on-policy RL into off-policy RL and proposed truncated importance sampling~\cite{yao}. Liu, Li et al.\ analyzed token- and sequence-level estimators and introduced masked importance sampling~\cite{liu}; IcePop masks and clips tokens with large discrepancy~\cite{icepop}. Marek and Ryabinin decomposed the off-policy gradient into a drift and a signal term and remove the drift by centering under $q$~\cite{centering}. Modal combined score centering, MIS and top-$p$ mask replay to train stably on samplers spanning six accelerator types, and argued that the growth of the mismatch, not its size, drives collapse~\cite{modal}. Decoupled PPO in AReaL~\cite{areal} and M2PO~\cite{m2po} can work with a recomputed $\pi_{\text{old}}$, but not without $q$. All of these weight the sampler's own tokens; substitution changes the token instead.

\paragraph{Bitwise alignment.} Batch-invariant kernels~\cite{thinkingmachines} and per-operator alignment of inference and training kernels~\cite{vllmbitwise} reach zero mismatch, at the price of running the same numerical path on both sides, which a permissionless network cannot impose.

\paragraph{Verification.} TOPLOC commits to activations and checks them with a tolerance~\cite{toploc}; INTELLECT-2 used it to admit untrusted rollouts~\cite{intellect2}. Neither checks the sampled token. Token-DiFR is the published form of a seed-synchronized check: given a shared seed, an honest provider's tokens match a reference more than 98\% of the time, which makes them auditable~\cite{difr}. Our match rate is the same check; the contribution is to reuse the seed to build the training target.

\paragraph{Couplings and counterfactuals.} Gumbel sampling is near-optimal among communication-free couplings and makes speculative decoding drafter-invariant~\cite{daliri}. Under a Gumbel-max structural causal model, the token another policy would have produced with the same noise is the counterfactual token~\cite{oberst}. Common random numbers (ARM, DisARM)~\cite{arm,disarm}, drafter-invariant speculative decoding~\cite{daliri} and Tree-Backup's replacement of the target action~\cite{treebackup} are the closest relatives. Each of them regenerates the continuation, evaluates the counterfactual reward or relies on a critic. Substitution does none of these and uses no probabilities from the sampler. To our knowledge, it has not been published.

\section{Limitations}\label{sec:limits}
\begin{enumerate}
\item \textbf{Biased gradient.} States and returns come from $q$, and cross-credit remains at disagreement positions. The branching test (Section~\ref{sec:stale}) suggests the bias is not concentrated on decision points, but covers math only.
\item \textbf{Residual drift as deployed.} The fp32 pick and bf16 score leave a small residual (Section~\ref{sec:notguaranteed}), measured on the last two checkpoints only. Admission gates add a residual bounded by the excluded fraction (about 2.5\% of rollouts), not measured directly.
\item \textbf{Adaptive samplers.} A sampler that uses the noise of future positions to choose its prefix, or withholds groups after seeing their draws, is outside the guarantee (Section~\ref{sec:threat}). Re-grading runs after the optimizer step, so a group whose reward changes is trained but not paid: an economic defense, not a training defense (70 production groups).
\item \textbf{Scope of the lab evidence.} The comparison of losses rests on a 2B model with one or two seeds, with L3 at $5\times10^{-6}$ on different hardware (H100) from L0 and L2 (A100), and long runs stopped before their planned 400 steps; none exists on the 9B model. With a realistic fp8 sampler over 120 steps, L2 trails L0 by about 0.05 reward. About 5--6\% of the gradient is rewritten for an honest sampler, and the drift and gradient-cosine measurements were made under inverse-CDF sampling only.
\item \textbf{Calibration and engine.} The 0.90 / 8 / 3 / 1 thresholds and the audit lag of 4 were calibrated at learning rate $2\times10^{-6}$ with simulated lag; they do not hold at $5\times10^{-6}$ on code. The draw relies on a pinned vLLM version and a non-fused float32 \texttt{logf}; speculative decoding is not supported.
\item \textbf{Multi-turn episodes.} No re-grading and no final end-of-sequence test; only the gap test applies. The cross-credit bias and the drift have not been measured at 32k--64k context or in multi-turn episodes; only the match rate has.
\item \textbf{Production run.} The run shows that the system trains, not that substitution beats an alternative: there is no arm without substitution on the same model and data, and at learning rate $2\times10^{-6}$ an uncorrected run also holds for a long time (in the lab it collapsed only at step 184), so the evidence that substitution prevents collapse comes from the lab (Section~\ref{sec:minitrain}). In addition, steps 1--16 did not train the head, embeddings or final norm; KL to the reference was not logged. The match rate declines on agentic tasks and instruction following (from 0.985 to 0.95--0.97) for a reason not yet established. Benchmark gains concentrate in the task families trained on.
\end{enumerate}

\section{Conclusion}
Public Gumbel noise turns the sampler's randomness into the trainer's. At each position the trainer scores the token its own policy would have drawn under that noise, in the sampler's context. The drift toward the sampler then vanishes in expectation for any hardware, precision and lag, without reading a single sampler probability, and the same comparison verifies and prices each sampler. In the lab, uncorrected training collapsed where substitution did not, and substitution performed comparably to a correction that needs the true $q$. On a permissionless network, 158 steps and 15.75B tokens from untrusted, heterogeneous samplers trained with no skipped step, 0.979 match rate and 1.21\% substitution, and the model improved on all eight held-out benchmarks, from 17.3 to 43.6 pass@1 on SWE-bench Verified. RL rollouts can therefore come from any GPU, owned by anyone, without trusting what the sampler reports. The main open question is a protocol that keeps the guarantee against samplers that adapt to future noise.


\appendix
\section{Engineering notes}\label{app:eng}
\subsection{A fairness bug: position desynchronization under preemption}\label{sec:preempt}
The first sampler implementation counted calls per slot. vLLM preemption reset the counter, so tokens were drawn with the wrong $u$. Desynchronization affected 47.5\% of honest rollouts on RTX 3090, 34.2\% on 5090, 28.9\% on L40 and 21.5\% on A6000: the smaller the card's KV cache, the more it was penalized. Indexing positions from the runner's token counts (Section~\ref{sec:noise}) gave 0 wrong positions out of 44.9M traced in the lab (31 runs, 4 cards, 9,489 preemption resumes), and 0 out of 6.8M after the port to production.

\subsection{Qualification of the production code}\label{sec:qualif}
\begin{itemize}
\item \textbf{8$\times$A100, IF, 9B SFT.} Match rate 0.992 and 0.35--0.5\% substitution at lag 0--1; match rate 0.991, 0.990, 0.978, 0.969 and 0.957 at lags 0 to 4. FSDP2 on 4 and 8 GPUs reproduces 1 GPU (cosine 0.99996). At learning rate $5\times10^{-6}$, an honest sampler fell below the floor because of lag, which led us to lower the floor to 0.90.
\item \textbf{End-to-end loops on one H100 (2B).} A first run failed 48 of 150 audits, all at an audit lag of 9 or more caused by queueing; audits now run newest first and are dropped beyond lag 4. A second run passed 172 of 173 audits with 0 groups rejected.
\item \textbf{8$\times$H100 rehearsal (final 9B SFT).} Match rate 0.9889--0.9926, substitution 0.44--0.74\%, 0 honest groups rejected; SWE and TMax groups audited and passed. Multi-turn EnvScaler: 25 of 25 groups passed, minimum per-rollout match rate 0.981.
\end{itemize}

\section{Gates, in order of application}\label{app:gates}
All constants are provisional.

\begin{table}[H]\centering\small
\begin{tabular}{rp{0.43\linewidth}lp{0.27\linewidth}}
\toprule
\# & Gate & Value & Consequence \\
\midrule
0 & Valid lease and $S_k$ not yet revealed & lag $\le$ 16 & Refused \\
1 & Coordinator-side grading and integrity; the reward is never supplied by the sampler & -- & Rejected, not paid \\
2 & Learning zone: standard deviation of the 16 rewards $\ge$ 0.24 & 0.24 & In zone: trained; out of zone: audited \\
3 & Lag above 16 checkpoints & 16 & Not trained, not paid \\
4 & Per rollout, in the training forward: match $\ge$ 0.90, max gap $\le$ 8, final EOS gap $\le$ 1 (not for episodes) & 0.90 / 8 / 1 & Removed from the loss, not paid \\
5 & Suspicious positions: gap $>$ 3, or $\vstar$ is a stop token and $\vstar \neq y$ & 3 & Feeds gate 8 \\
6 & Group: at least one admitted rollout, pooled match $\ge$ 0.90, no rollout $<$ 0.5 & 0.90 / 0.5 & Rejected: rollouts cancelled by a second pass with negated advantages \\
7 & Step health: gradient norm $>$ 100 or non-finite & 100 & Step skipped \\
8 & Re-grading with the trainer's picks at suspicious positions (truncated at a stop-token pick) & -- & Reward changed: not paid, already trained \\
9 & Out-of-zone audits: 15\%, or 100\% on probation, against the latest checkpoint, dropped beyond lag 4 & 15\% / 4 & Failure: not paid, back to probation \\
\bottomrule
\end{tabular}
\end{table}
Lag has no quality ceiling of its own: it lowers the match rate, which the floors judge.

\section{Training hyperparameters (production)}
\begin{table}[H]\centering\small
\begin{tabular}{ll}
\toprule
Parameter & Value \\
\midrule
Model & 9B (Teutonic-I, after SFT) \\
Sampling & Temperature 1, top-$p$ 1, top-$k$ 0, Gumbel-max \\
Rollouts per group / groups per step & 16 / 512 \\
Optimizer & AdamW, fp32 master weights, 8-bit states \\
Learning rate & $2\times10^{-6}$, 8 warm-up steps \\
Gradient clipping & Norm 1.0; step skipped above 100 \\
KL coefficient & 0 \\
Checkpoint publication & Every optimizer step \\
Overlong penalty & 0.1 ramp over the last 4,096 tokens \\
\bottomrule
\end{tabular}
\end{table}

\section{Checkpoint evaluations during the run}\label{app:ckpt}
Run-time protocol (Section~\ref{sec:bench}): temperature 1, top-$p$ 1, top-$k$ 0; AIME 30 problems $\times$ 8 samples, GPQA Diamond 198 $\times$ 4, LiveCodeBench 300 $\times$ 2, IFBench 300 $\times$ 2, MMLU-Pro 500 $\times$ 2, SWE-bench Verified 100 $\times$ 2, TMax 200 $\times$ 4. pass@1 (\%), 95\% bootstrap intervals over problems at checkpoints 0 and 85.

\begin{table}[H]\centering\small
\setlength{\tabcolsep}{3.5pt}
\resizebox{\linewidth}{!}{%
\begin{tabular}{lccccccccc}
\toprule
Set & 0 & 20 & 40 & 60 & 85 & 100 & 120 & 140 & 158 \\
\midrule
AIME 2025 & 75.8 {\scriptsize[65.0--85.4]} & 78.8 & 86.2 & 83.3 & 87.1 {\scriptsize[77.1--95.4]} & 87.9 & 86.2 & 90.8 & 90.4 \\
AIME 2026 & 73.3 {\scriptsize[60.8--85.4]} & 80.4 & 81.7 & 81.7 & 84.6 {\scriptsize[74.2--92.9]} & 86.7 & 87.1 & 90.8 & 89.6 \\
GPQA Diamond & 62.8 {\scriptsize[57.6--68.1]} & 65.5 & 66.0 & 65.2 & 65.0 {\scriptsize[59.3--70.3]} & 65.7 & 65.3 & 65.0 & 65.9 \\
LiveCodeBench & 39.5 {\scriptsize[34.5--44.5]} & 41.8 & 49.5 & 49.0 & 55.0 {\scriptsize[49.5--60.2]} & 58.7 & 56.5 & 59.3 & 61.0 \\
IFBench & 63.7 {\scriptsize[58.8--68.2]} & 62.8 & 68.5 & 68.3 & 73.0 {\scriptsize[68.8--77.0]} & 70.2 & 71.3 & 72.5 & 74.7 \\
MMLU-Pro & 75.0 {\scriptsize[71.4--78.5]} & 74.5 & 75.8 & 76.0 & 77.9 {\scriptsize[74.5--81.1]} & 76.8 & 77.9 & 79.1 & 78.4 \\
SWE-bench V.\ (100) & 12.0 {\scriptsize[7.0--18.0]} & 16.0 & 25.0 & 29.5 & 38.5 {\scriptsize[30.0--47.0]} & 44.0 & 45.0 & 43.5 & 48.5 \\
TMax & 17.2 {\scriptsize[13.5--21.0]} & 21.9 & 37.0 & 48.0 & 61.3 {\scriptsize[55.8--66.5]} & 67.1 & 69.8 & 72.0 & 74.6 \\
\bottomrule
\end{tabular}
}
\caption{Run-time checkpoint evaluations. Checkpoint 0 is the SFT model. TOPLOC verification in addition to the draw for checkpoints 0--140.}
\end{table}


\begin{thebibliography}{99}\small
\bibitem{intellect2} Prime Intellect Team (S.~Jaghouar et al.). INTELLECT-2: A reasoning model trained through globally decentralized reinforcement learning. arXiv:2505.07291, 2025.
\bibitem{beam} Reflection AI. Introducing Beam: Reflection's 501B open-weight model. \url{https://reflection.ai/blog/introducing-beam}, October 5, 2026.
\bibitem{mai} The Microsoft AI Team. MAI-Thinking-1: Building a hill-climbing machine. Technical report, Microsoft AI, 2026, p.~51. \url{https://microsoft.ai/pdf/mai-thinking-1.pdf}.
\bibitem{modal} N.~Jiang. Toward RL on mixed-precision, mixed-accelerator samplers. Modal, \url{https://modal.com/blog/toward-rl-on-mixed-precision-mixed-accelerator-samplers}, October 13, 2026.
\bibitem{yao} F.~Yao, L.~Liu, D.~Zhang, C.~Dong, J.~Shang, J.~Gao. Your efficient RL framework secretly brings you off-policy RL training, 2025; On the rollout-training mismatch in modern RL systems, NeurIPS 2025.
\bibitem{liu} J.~Liu, Y.~Li, Y.~Fu, J.~Wang, Q.~Liu, Z.~Jiang. When speed kills stability: Demystifying RL collapse from the training-inference mismatch, 2025.
\bibitem{icepop} Ling Team. Every step evolves: Scaling reinforcement learning for trillion-scale thinking model (Ring-1T, IcePop). arXiv:2510.18855, 2025.
\bibitem{centering} M.~Marek, M.~Ryabinin. Score centering stabilizes off-policy reinforcement learning. arXiv:2609.20807, 2026.
\bibitem{difr} A.~Karvonen, D.~Reuter, R.~Rinberg, L.~Marks, A.~Garriga-Alonso, K.~Warr. DiFR: Inference verification despite nondeterminism. Second Workshop on Technical AI Governance Research at ICML 2026; arXiv:2511.20621, 2025.
\bibitem{daliri} M.~Daliri, C.~Musco, A.~T.~Suresh. Coupling without communication and drafter-invariant speculative decoding. arXiv:2408.07978, 2024.
\bibitem{oberst} M.~Oberst, D.~Sontag. Counterfactual off-policy evaluation with Gumbel-max structural causal models. ICML, 2019.
\bibitem{toploc} J.~M.~Ong et al. TOPLOC: A locality sensitive hashing scheme for trustless verifiable inference. ICML, 2025.
\bibitem{thinkingmachines} H.~He, Thinking Machines Lab. Defeating nondeterminism in LLM inference. \url{https://thinkingmachines.ai/blog/defeating-nondeterminism-in-llm-inference/}, September 10, 2025.
\bibitem{vllmbitwise} Y.~Bai, A.~Shen. Finish RL training faster: Bitwise-consistent, high-performance, agent-driven. vLLM blog, \url{https://vllm.ai/blog/2026-10-14-train-inference-consistency}, October 14, 2026.
\bibitem{gumbel} E.~J.~Gumbel. Statistical theory of extreme values and some practical applications. National Bureau of Standards, Applied Mathematics Series 33, 1954.
\bibitem{maddison} C.~J.~Maddison, D.~Tarlow, T.~Minka. A* sampling. NeurIPS, 2014.
\bibitem{grpo} Z.~Shao et al. DeepSeekMath: Pushing the limits of mathematical reasoning in open language models. arXiv:2402.03300, 2024.
\bibitem{ppo} J.~Schulman, F.~Wolski, P.~Dhariwal, A.~Radford, O.~Klimov. Proximal policy optimization algorithms. arXiv:1707.06347, 2017.
\bibitem{vllm} W.~Kwon et al. Efficient memory management for large language model serving with PagedAttention. SOSP, 2023.
\bibitem{dapo} Q.~Yu et al. DAPO: An open-source LLM reinforcement learning system at scale. arXiv:2503.14476, 2025.
\bibitem{gspo} C.~Zheng et al. Group sequence policy optimization. arXiv:2507.18071, 2025.
\bibitem{cispo} MiniMax. MiniMax-M1: Scaling test-time compute efficiently with lightning attention (CISPO). arXiv:2506.13585, 2025.
\bibitem{areal} W.~Fu et al. AReaL: A large-scale asynchronous reinforcement learning system for language reasoning. arXiv:2505.24298, 2025.
\bibitem{m2po} H.~Zheng, J.~Zhao, B.~Chen. Prosperity before collapse: How far can off-policy RL reach with stale data on LLMs? arXiv:2510.01161, 2025.
\bibitem{arm} M.~Yin, M.~Zhou. ARM: Augment-REINFORCE-merge gradient for stochastic binary networks. ICLR, 2019.
\bibitem{disarm} Z.~Dong, A.~Mnih, G.~Tucker. DisARM: An antithetic gradient estimator for binary latent variables. NeurIPS, 2020.
\bibitem{treebackup} D.~Precup, R.~S.~Sutton, S.~Singh. Eligibility traces for off-policy policy evaluation. ICML, 2000.
\bibitem{gpqa} D.~Rein et al. GPQA: A graduate-level Google-proof Q\&A benchmark. arXiv:2311.12022, 2023.
\bibitem{lcb} N.~Jain et al. LiveCodeBench: Holistic and contamination free evaluation of large language models for code. arXiv:2403.07974, 2024.
\bibitem{ifbench} V.~Pyatkin et al. Generalizing verifiable instruction following. arXiv:2507.02833, 2025.
\bibitem{mmlupro} Y.~Wang et al. MMLU-Pro: A more robust and challenging multi-task language understanding benchmark. arXiv:2406.01574, 2024.
\bibitem{swebench} C.~E.~Jimenez et al. SWE-bench: Can language models resolve real-world GitHub issues? arXiv:2310.06770, 2023.
\bibitem{tmax} H.~Ivison, J.~O.~Yin, R.~Shao, T.~Xiao, N.~Lambert, H.~Hajishirzi. Tmax: A simple recipe for terminal agents. arXiv:2606.23321, 2026. Dataset: \url{https://huggingface.co/datasets/allenai/TMax-15K}.
\bibitem{sweagent} J.~Yang et al. SWE-agent: Agent-computer interfaces enable automated software engineering. NeurIPS, 2024; arXiv:2405.15793. mini-swe-agent: \url{https://github.com/SWE-agent/mini-swe-agent}.
\end{thebibliography}
\end{document}